\documentclass[journal]{IEEEtran}
\usepackage{amsmath}
\usepackage{amssymb}
\usepackage{url}
\usepackage{graphicx}
\usepackage{booktabs}
\usepackage{multirow}

\UseRawInputEncoding
\begin{document}

\title{An optimization method for out-of-distribution anomaly detection models}

\author{Ji Qiu, Hongmei Shi*, Yu Hen Hu, and Zujun Yu
\thanks{This work was supported by China Energy Group research project under Grant GJNY-21-65.}
\thanks{J. Qiu, H. Shi, and Z. Yu are with Beijing Jiaotong University, Beijing, 100044, P.R.China (e-mail: qiuji@bjtu.edu.cn; hmshi@bjtu.edu.cn; zjyu@bjtu.edu.cn)}
\thanks{Y.H. Hu is with University of Wisconsin-Madison, Madison, WI 53705, USA (e-mail: yhhu@wisc.edu).}}

\markboth{Journal of \LaTeX\ Class Files, Vol. 00, No. 00, Jan 2023}
{Shell \MakeLowercase{\textit{et al.}}: Bare Demo of IEEEtran.cls for IEEE Journals}
\maketitle

\begin{abstract}
Frequent false alarms impede the promotion of unsupervised anomaly detection algorithms in industrial applications. Potential characteristics of false alarms depending on the trained detector are revealed by investigating density probability distributions of prediction scores in the out-of-distribution anomaly detection tasks. An SVM-based classifier is exploited as a  post-processing module to identify false alarms from the anomaly map at the object level. Besides, a sample synthesis strategy is devised to incorporate fuzzy prior knowledge on the specific application in the anomaly-free training dataset. Experimental results illustrate that the proposed method comprehensively improves the performances of two segmentation models at both image and pixel levels on two industrial applications.
\end{abstract}


\IEEEpeerreviewmaketitle

\section{Introduction}

\IEEEPARstart{D}{ue} to the long-tail distribution of anomaly classes and the high cost of professional annotation, promising methods transform the anomaly detection task into an out-of-distribution(OOD) \cite{OOD1, OOD2, OOD3} problem. Unsupervised anomaly detection models learn from an anomaly-free dataset and measure the distance between a test image and the learned distribution \cite{AD1, AD2, AD3}. The granularities vary from image-level classification to pixel-level segmentation depending on the requirement of a specific application. Please refer to \cite{IEEEReview} for a comprehensive survey of anomaly detection methods.

\begin{figure}[]
\centering
\includegraphics[width=0.9\linewidth]{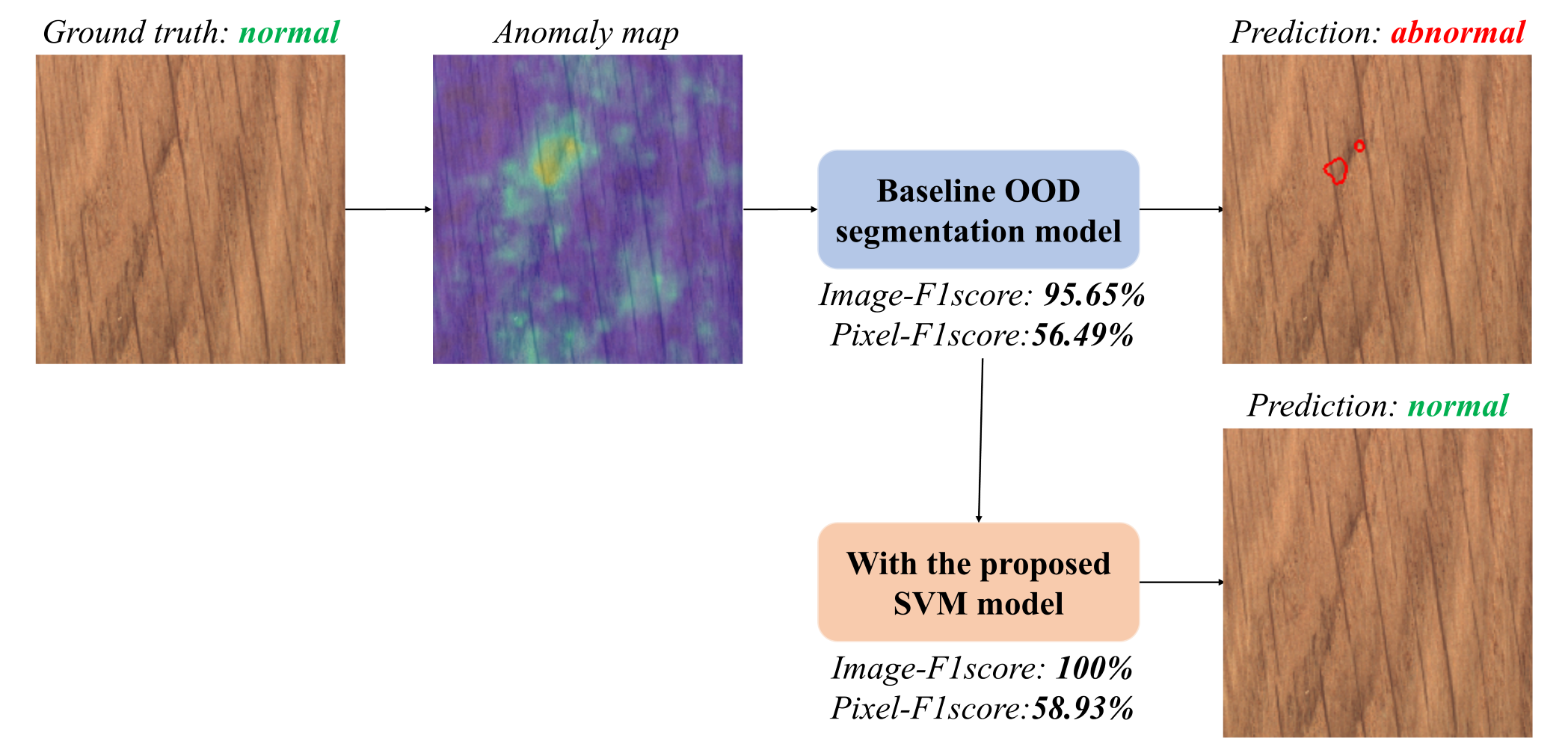}
\caption{Detection results of an OOD model without and with the proposed optimization method on a wood defect detection task. The baseline segmentation model is Fastflow \cite{fastflow}, consisting of a ResNet-18 backbone \cite{ResNet} initialized with the ImageNet pre-trained weights and a normalizing flow network trained by the corresponding anomaly-free dataset in the MVTec AD dataset \cite{MVTec AD}. Parameters of baseline segmentation models freeze during the testing process.}
\label{Fig: woodimprove}
\end{figure}

In this Letter, we mainly focus on detection models that generate prediction scores synthetically at image and pixel levels as shown in Fig. \ref{Fig: woodimprove}. The image-level prediction score indicates if defects exist in the corresponding image (abnormal/normal). Meanwhile, pixel-level prediction scores compose the anomaly map that will be converted into the pixel-wise segmentation mask by thresholding.

Frequent false alarms impede the promotion of anomaly detectors in industrial defect inspection applications \cite{FA1, FA2, FA3}. By investigating concrete false-alarm regions, we find that these regions share a different appearance from common patterns in the anomaly-free training dataset. The definition of defects in industrial applications is more complex than naive global or local appearance differences, resulting in a diversity of intra-class features in the training dataset. Unsupervised models are intrinsically prone to overfitting on common patterns since the absence of defect samples. Inadequate learning of few-sample patterns leads to false positives of similar ones in the testing stage. Hence, noise and complexity in practical applications aggravate the overfitting problem and cause inferior performance.

Due to the rarity of anomalies, a low-probability false alarm rate results in a high proportion of false positives in outputs and causes adverse effects on the employment of unsupervised defect detection models in industrial applications. To this end, we provide an optimization approach to filter out false alarms and leverage the detection performance. In summary, our contributions are:

(1) A post-processing optimization method is proposed to identify false alarms from OOD anomaly detection models through an SVM classifier at the object level according to fuzzy knowledge of the specific application.

(2) A training sample synthetic strategy for the SVM classifier is devised to investigate model-dependent samples from existing anomaly maps on the anomaly-free training dataset.

(3) Experimental results on two industrial applications illustrate that the proposed method comprehensively improves the detection performances of two state-of-the-art unsupervised anomaly detectors.

\section{Proposed Method}

\subsection{False alarm in Unsupervised Anomaly Detection}

Unsupervised anomaly detection falls into the OOD domain as detectors learn from the anomaly-free training dataset. Given an anomaly-free training dataset, $\chi = \left \{ x_{0},x_{1},...,x_{D} \right \} $ with $x_{i} \in \mathbb{R}^{M}$, the OOD model aims to train a neural network-enabled anomaly score learning function $\tau: \chi \to K$ that differentiates the anomalies from anomaly-free ones. The separation unit is an image, object, or pixel, depending on the granularity of the specific visual task. To investigate the cause of false alarms in OOD defect segmentation tasks, we may have a further statistical analysis of the pixel-wise transformed outputs $\tau(\chi)$, which are prediction scores on anomaly maps of $\chi$. Generally, the score falls in $[0,1]$ while a higher value indicates a higher likelihood of belonging to a defect.

\begin{figure}[ht]
\centering
    \begin{minipage}{0.32\linewidth}
		\centering\includegraphics[width=1.0\linewidth]{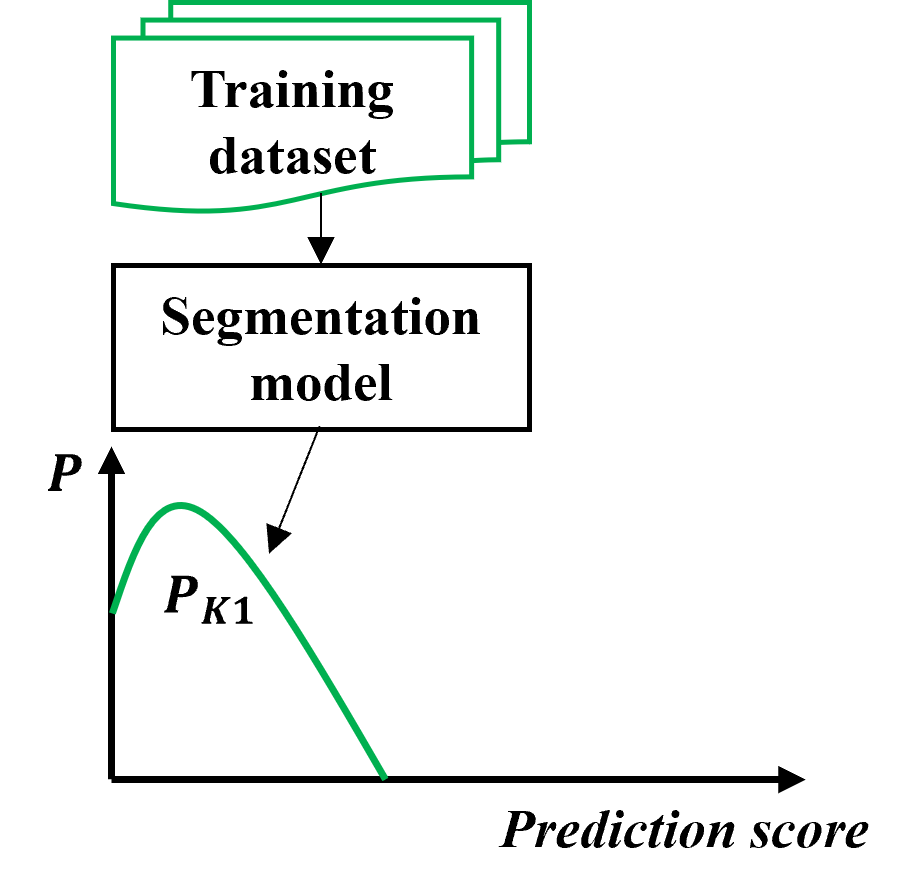}
        \centerline{(a)}
	\end{minipage}
    \begin{minipage}{0.32\linewidth}
		\centering\includegraphics[width=1.0\linewidth]{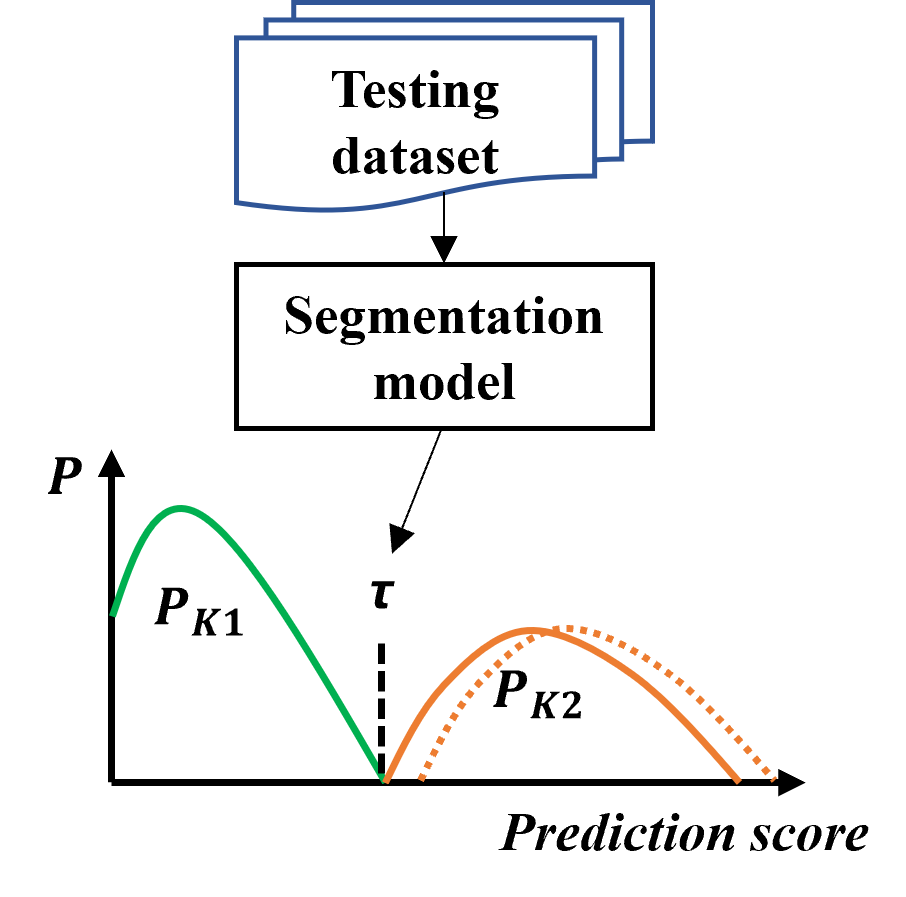}
        \centerline{(b)}
	\end{minipage}
    \begin{minipage}{0.32\linewidth}
		\centering\includegraphics[width=1.0\linewidth]{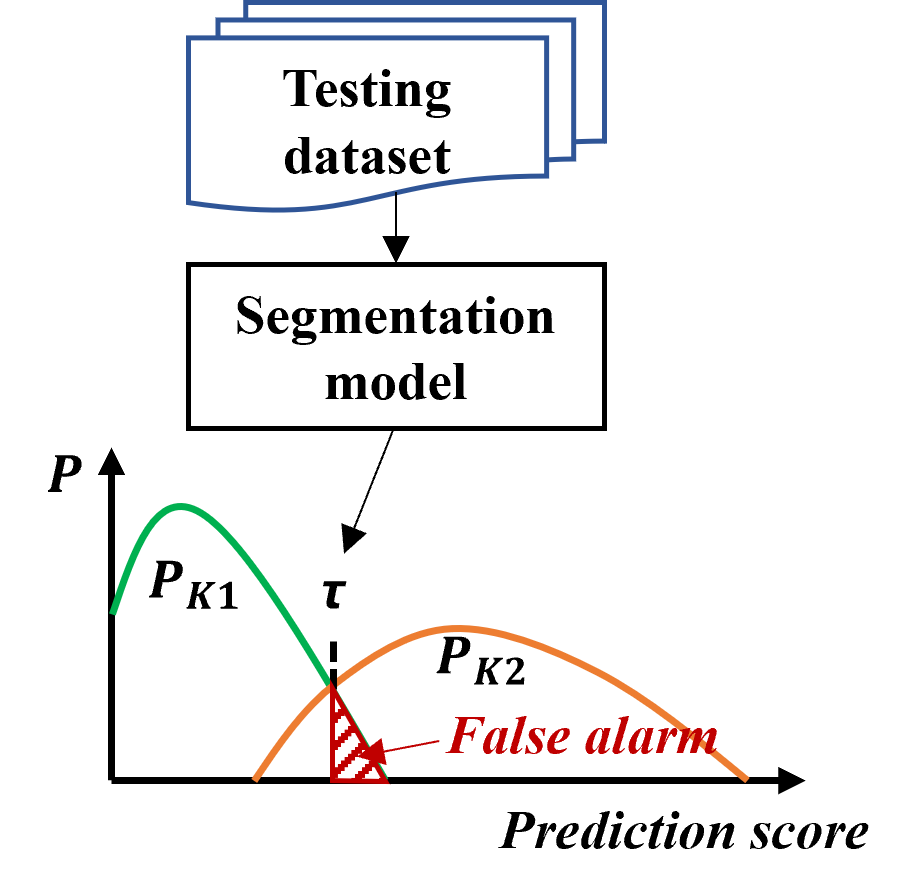}
        \centerline{(c)}
	\end{minipage}
\caption{Density probability distributions of prediction scores. (a) depicts the distribution of anomaly-free targets $P_{K1}$ produced by the segmentation model in the training process. (b) shows the ideal condition of OOD models: When the distribution of anomaly ones $P_{K2}$ forms as the solid orange line, one threshold exists for successful detection with a perfect AUROC score. Similarly, more thresholds exist for distant distributions like the dotted orange line. (c) presents an actual condition that distributions intersect due to inferior discrimination ability. False alarms rise and lay in the red region.}
\label{Fig: Distribution}
\end{figure}

Unsupervised detectors are prone to overfit common patterns in the training dataset as the absence of defects from the training dataset. In practical applications, anomaly-free images involve various adverse factors causing appearance differences, including lighting conditions, clutter background, product batch, slight in-service aging, and handwriting stamps. At the same time, defects are complex with long-tail distribution classes. Some anomalies might be significant color differences covering the main body, while some might be small shape changes concerning only a few pixels. Discrimination abilities of unsupervised detectors are usually incompatible with the complex classification boundary between defect and defect-free in noisy and complex industrial applications.

As shown in Fig. \ref{Fig: Distribution}(c), while two distributions intersect, the resulting performance will have deficiencies no matter where the threshold $\tau$ locates. As the positive in defect detection tasks refers to the defect, those defect-free pixels with higher scores than the threshold become false alarms. Simply increasing $\tau$ to avoid the intersection with $P_{K1}$ may eliminate false alarms while causing ignorance of positive defects. Therefore, the false-alarm problem comes from the inadequate discrimination ability to differentiate $P_{K1}$ and $P_{K2}$. The performance can hardly optimize through existing parameter modification since there is no information on the out-of-distribution. To this end, we provide a post-processing optimization approach to incorporate additional fuzzy knowledge of defects within the context of specific applications that automatically eliminates false alarms while preserving the unsupervised learning architecture.

\subsection{Proposed post-processing optimization method}

\begin{figure}
\centering
\includegraphics[width=1.0\linewidth]{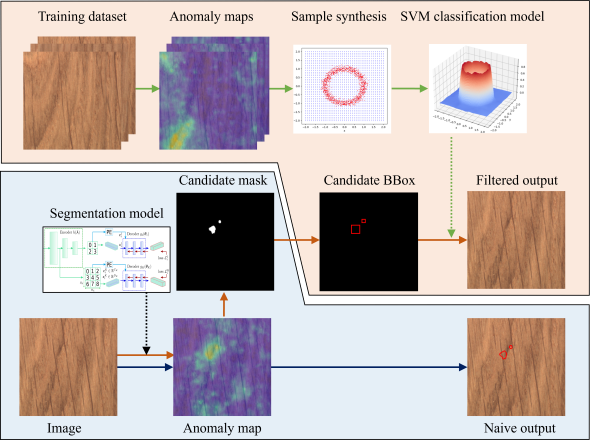}
\caption{The proposed post-processing optimization workflow. Blue arrows describe the baseline defect detection processes that directly generate outputs from anomaly maps. Orange arrows present flows of the devised method, which takes the generation of the anomaly map from the segmentation model as the first stage and filters out object-level false alarms on anomaly maps according to fuzzy domain knowledge at the second stage. Green arrows draw the sample synthesis and model training process of the classification model.}
\label{Fig: framework}
\end{figure}

Fig. \ref{Fig: framework} depicts the optimization process of the proposed unsupervised defect detection method, where the proposed SVM classifier connects after the baseline defect detection model to filter out false alarms.

In the output process of a baseline defect detection algorithm, the naive output, a segmentation mask, comes directly from the anomaly map via a thresholding operation. Since false alarms depend on the detection model and exist even in the anomaly-free training dataset, we may train a classification model to learn their model-dependent object-level features in anomaly maps of existing images. Object-level features are physical constraints, including size, color, shape, and distribution. They are discriminative, intuitive, and flexible to incorporate fuzzy knowledge within the context of specific applications. As shown in red arrows, the proposed classifier filters out false alarms from candidate defects exploiting the joint distribution of size and scale in the following experiments.

The classification model is a soft margin SVM classifier that describes fuzzy domain knowledge on physical constraints. Donate the input vector as $o_{n}$, the label indicating a false alarm or a defect as $y_{n}$, and the distance between the closest sample to the hyperplane is:

\begin{align}
margin(w) = \min_{n = 1,...,N}distance(o_{n},w)
\end{align}

Hence, the objective function of the classifier is:

\begin{align}
\begin{split}
&\max_{w}  margin(w)\\
&\text{subject to } y_{n}w^{T}x_{n}> 0
\end{split}
\end{align}

The kernel of feature transformation is the radial basis function,  a.k.a Gaussian kernel, which improves the computation efficiency of this quadratic programming problem. Kernel learning maps the original linear indivisibility distributions into separable higher dimensions with limited computation cost. The original input space consistent with $n$ input samples may turn into the separable feature dimension with a maximum of $n$ dimensions through transformation. Donating the mapping function as $\phi(\cdot)$ and width of the kernel as $\sigma$,the radial basis function is:

\begin{align}
K(o_{i},o_{j}) = \phi(o_{i}) \cdot \phi(o_{j}) = exp(-\frac{(\left\| o_{i}-o_{j} \right\|)^{2}}{2{\sigma}^{2}})
\end{align}

In the testing stage, $o_{n}$ automatically comes from bounding boxes of object-level defect candidates on the anomaly map. The classification allocates a false-alarm probability score for each candidate, which determines post-processes at pixel and image levels. On the one hand, pixels belonging to high-score regions will multiply by a positive coefficient to shrink their effects on the anomaly map, which contributes to eliminating false alarms on the segmentation mask. On the other hand, the image-level prediction score of an image descends according to the area ratio of high-score regions and the overall image-level prediction score distribution.

\subsection{Unsupervised Sample synthesis for classifier training}

A dilemma of the proposed method is training the classification model without balanced samples. To provide training samples for the SVM classifier, we devise a sample synthesis strategy that generates binary class samples depending on the specific detector and application.

\begin{figure}
\centering
\includegraphics[width=1.0\linewidth]{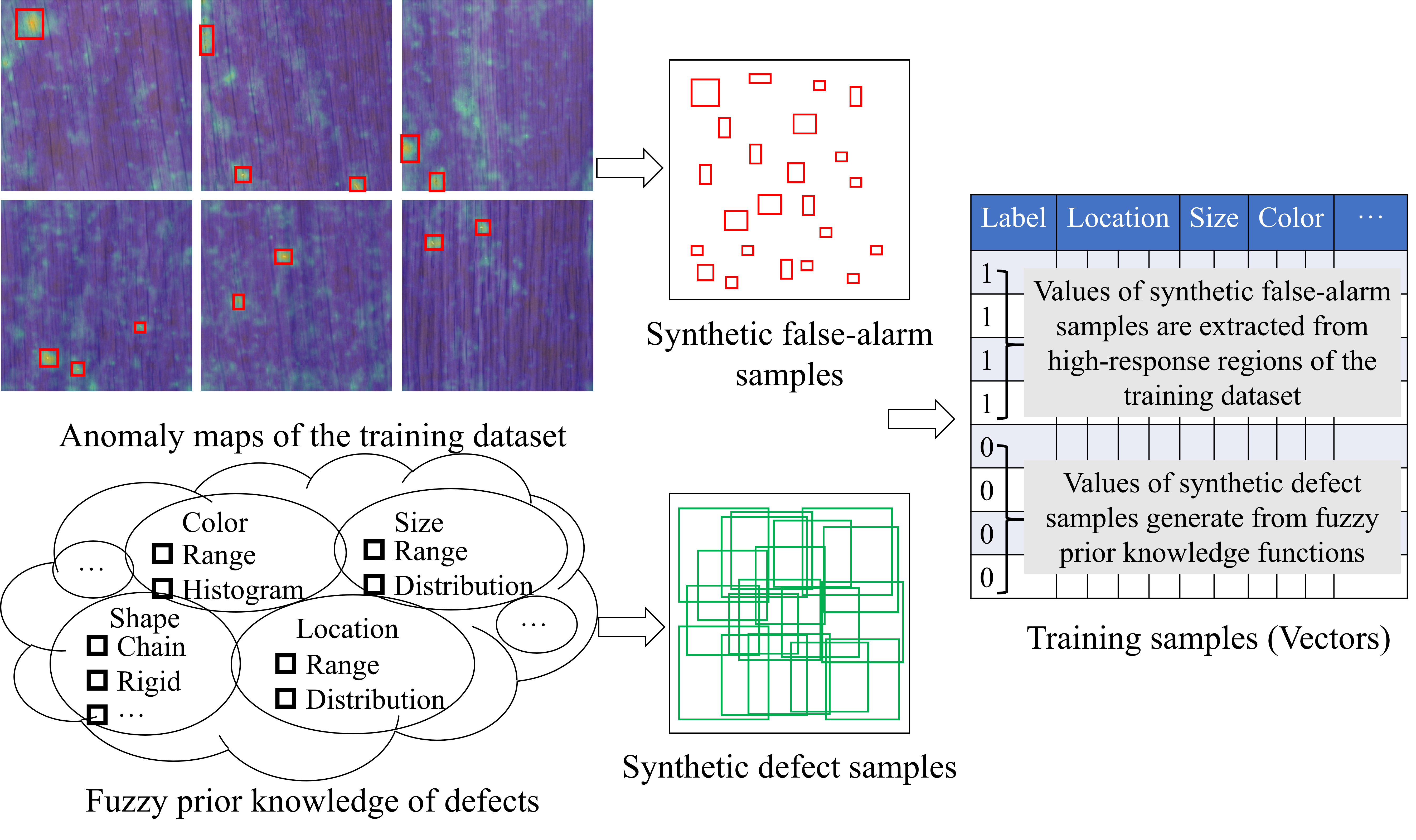}
\caption{Sample synthesis workflow. Training samples for the classifier are in the form of vectors, whose dimension depends on selected discriminative prior knowledge descriptions. Synthetic defect samples generate from fuzzy knowledge, while synthetic false-alarm ones come from high-response regions of the anomaly-free training dataset.}
\label{Fig: anomaps}
\end{figure}

As fuzzy prior knowledge provides physical constraints at the object level,  the dimension of classifier training samples (vectors) is determined by discriminative attributes of the specific application. As shown in Fig. \ref{Fig: anomaps}, dimensions of training samples are semantic elements. Synthetic defect samples generate directly from fuzzy knowledge, which summarizes as joint or independent descriptions of size, color, shape, location, and other aspects. For example, defects might be a rigid square shape with sizes ranging in $[10,18]$ and only appear in the central area of the image. These descriptions transform into a defect sample generator for this application that randomly produces defect samples (vectors) with joint constraints on size and location coordinates.

To obtain information on false alarms, we investigate anomaly maps of the training dataset. The anomaly map ($H*W*1$) is an intermediate product of the pixel-wise prediction, which reflects estimations of the detector for pixel-wise anomaly degrees of the image ($H*W*3$). Different from Fig. \ref{Fig: Distribution}(b), anomaly maps of the training samples as shown in Fig. \ref{Fig: anomaps} contain high response regions when the detector has inferior performance on a complex application. As pixel-wise prediction comes from the thresholding of the anomaly map, these yellow or green color areas in Fig.\ref{Fig: anomaps} will be predicted as defects. This observation reveals the inadequate discrimination ability of the detector on the training dataset, which also leads to false alarms in the testing stage. Similar to the test stage, regions made up of high-score pixels are detector-dependent false alarms in the anomaly-free training dataset. Taking these high-response patches as object-level false alarms, we may extract the corresponding dimensions (attributes) of synthetic false-alarm samples (vectors).

Besides, synthesis defect samples share a considerable quantity with false-alarm ones to balance the sample distribution of the SVM classifier. Basic augmentation strategies, including noise adding, position translation, and symmetry, are adopted in light of the specific application.

\section{Experiments}

\subsection{Experimental settings}

We conduct experiments on two industrial applications: wood defect examination (the MVTec AD dataset\cite{MVTec AD}: wood) and freight train monitoring (TFDS-RP dataset). Two SOTA defect detection algorithms, Fastflow \cite{fastflow} and Cflow \cite{cflow}, are chosen as baseline models, whose performances are compared with revised ones with the proposed post-processing model.

\subsection{Quantitative experimental results}

Quantitative results in Table \ref{tab: wood} and Table \ref{tab: TFDS} compare AUROC and F1-score at image and pixel levels, which validate comprehensive enhancements on baseline algorithms.

\begin{table}[ht]
\centering
\caption{quantitative performances on MVTec-wood dataset}
\label{tab: wood}{
\begin{tabular}{@{}ccccc@{}}
\toprule
\multirow{2}{*}{Model} & \multicolumn{2}{l}{Image-level metrics} & \multicolumn{2}{l}{Pixel-level metrics} \\ \cmidrule(l){2-5}
                & AUROC & F1-score & AUROC  & F1-score \\ \midrule
Fastflow        & 1.0000   & 0.9565   & 0.9698 & 0.5649   \\
Filter Fastflow & 1.0000   & 1.0000   & 0.9828 & 0.5893   \\
Cflow           & 1.0000   & 0.9167   & 0.9741 & 0.5811   \\
Filtered Cflow  & 1.0000   & 1.0000   & 0.9863 & 0.6208   \\\bottomrule
\end{tabular}
}
\end{table}
\begin{table}[ht]
\centering
\caption{quantitative performances on TFDS-RP dataset}
\label{tab: TFDS}{
\begin{tabular}{@{}ccccc@{}}
\toprule
\multirow{2}{*}{Model} & \multicolumn{2}{l}{Image-level metrics} & \multicolumn{2}{l}{Pixel-level metrics} \\ \cmidrule(l){2-5}
                & AUROC & F1-score & AUROC  & F1-score \\ \midrule
Fastflow        & 0.7236   & 0.2857   & 0.9113 & 0.0785   \\
Filter Fastflow & 0.9474   & 0.4000   & 0.9136 & 0.1524   \\
Cflow           & 0.6578   & 0.4615   & 0.8864 & 0.0684   \\
Filtered Cflow  & 0.8026   & 0.7500   & 0.9341 & 0.1332   \\\bottomrule
\end{tabular}
}
\end{table}

\subsection{Visual experimental results for pixel-wise segmentation}

\begin{figure}[ht]
\centering
\includegraphics[width=1.0\linewidth]{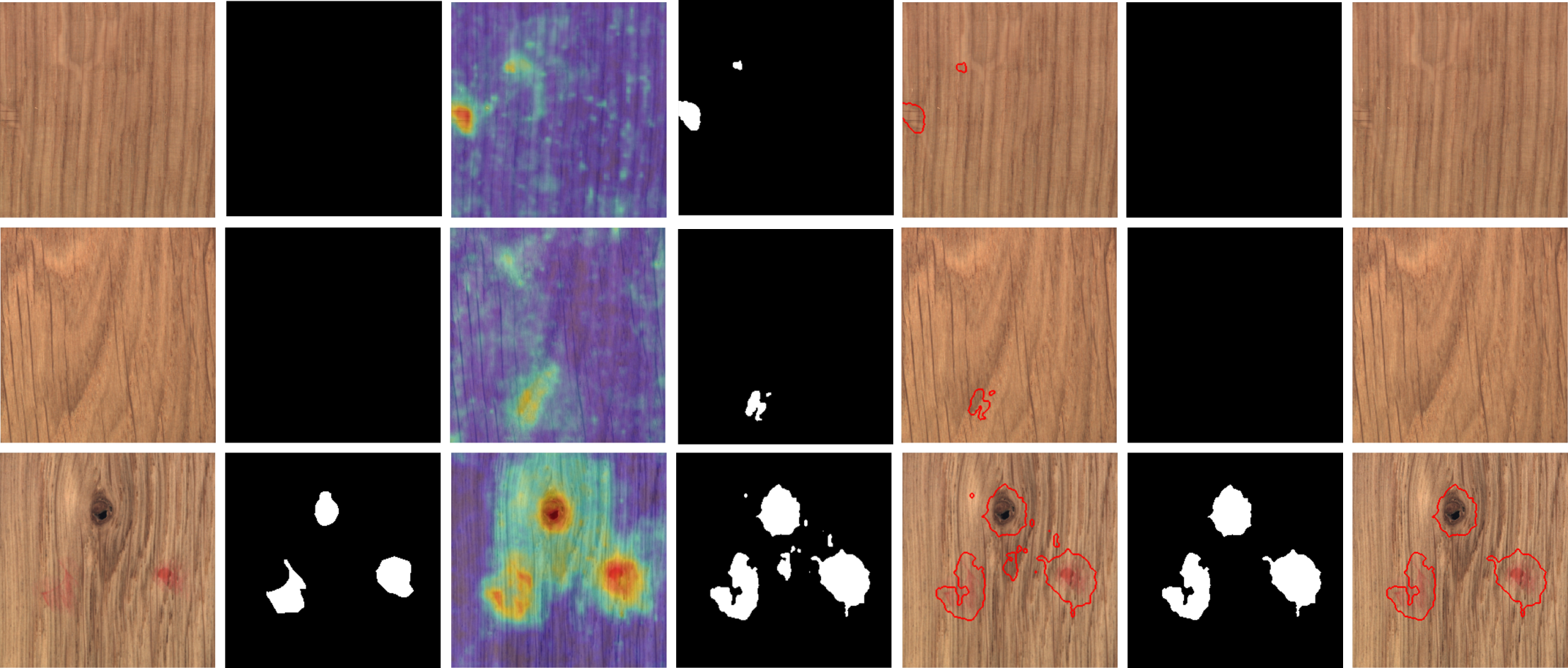}
\caption{Visualization of comparative experiments on wood defect examination using Fastflow. Columns correspond to the test image, ground truth, anomaly map, baseline defect mask, baseline segmentation result, filtered mask, and filtered segmentation result.}
\label{Fig: Fastflowwood}
\end{figure}

\begin{figure}[ht]
\centering
\includegraphics[width=1.0\linewidth]{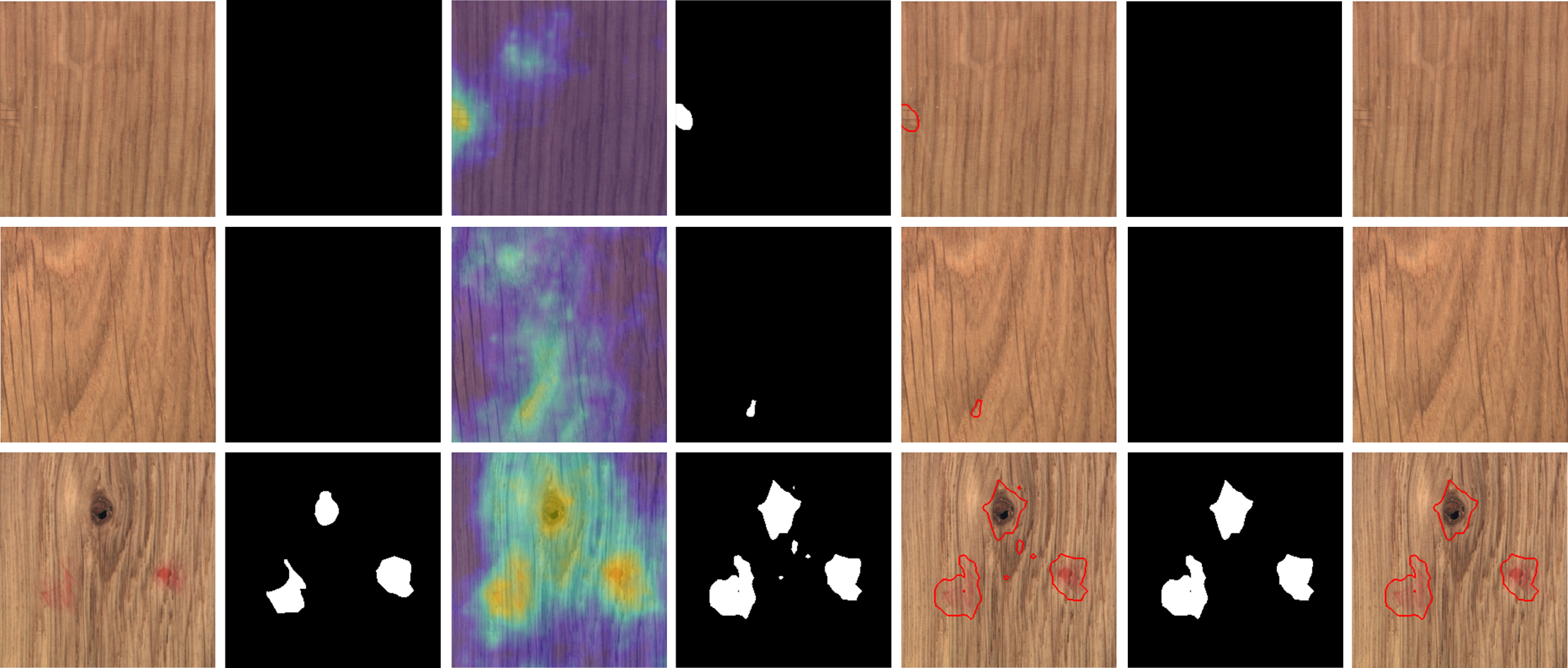}
\caption{Visualization of comparison experiments on wood defect examination using Cflow. Column settings are identical to Fig. \ref{Fig: Fastflowwood}.}
\label{Fig: Cflowwood}
\end{figure}

\begin{figure}[ht]
\centering
\includegraphics[width=1.0\linewidth]{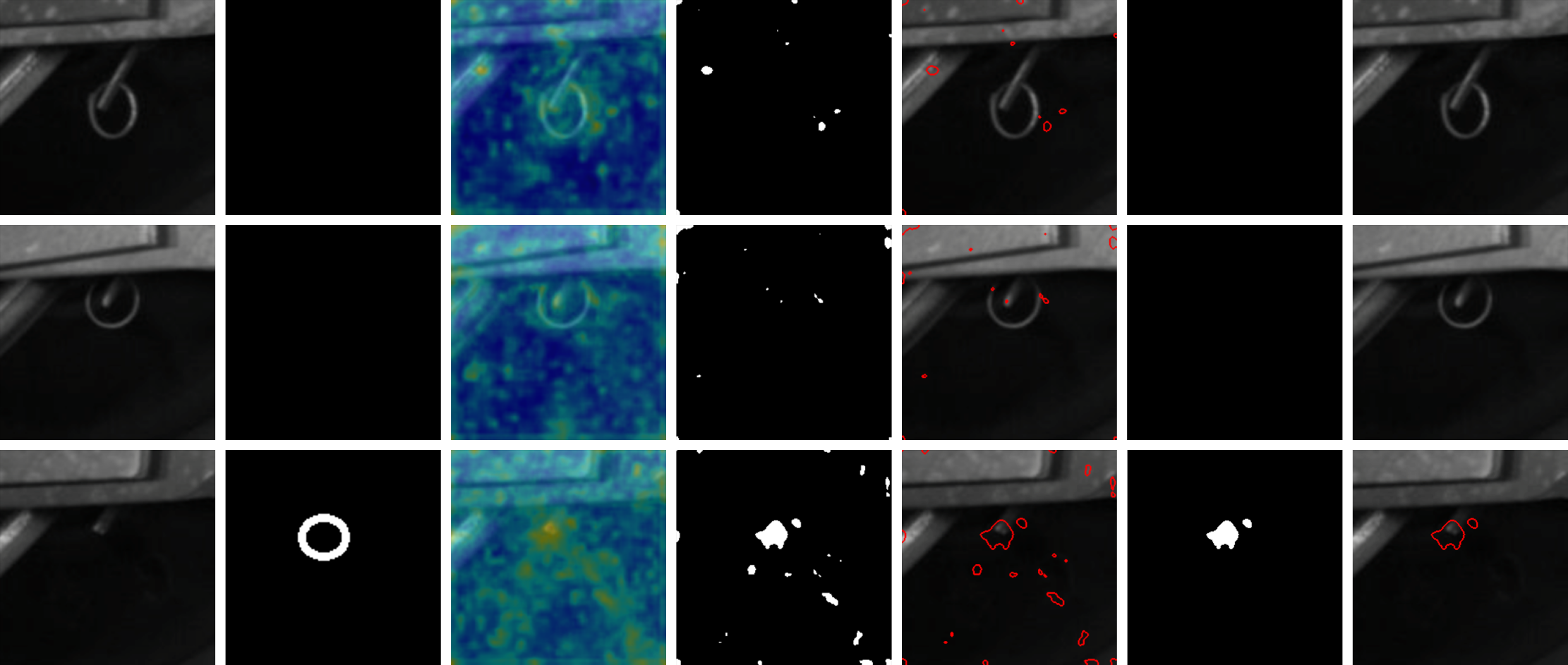}
\caption{Visualization of comparison experiments on round pin examination of freight train using Fastflow. Column settings are identical to Fig. \ref{Fig: Fastflowwood}.}
\label{Fig: FastflowTFDS}
\end{figure}

\begin{figure}[ht]
\centering
\includegraphics[width=1.0\linewidth]{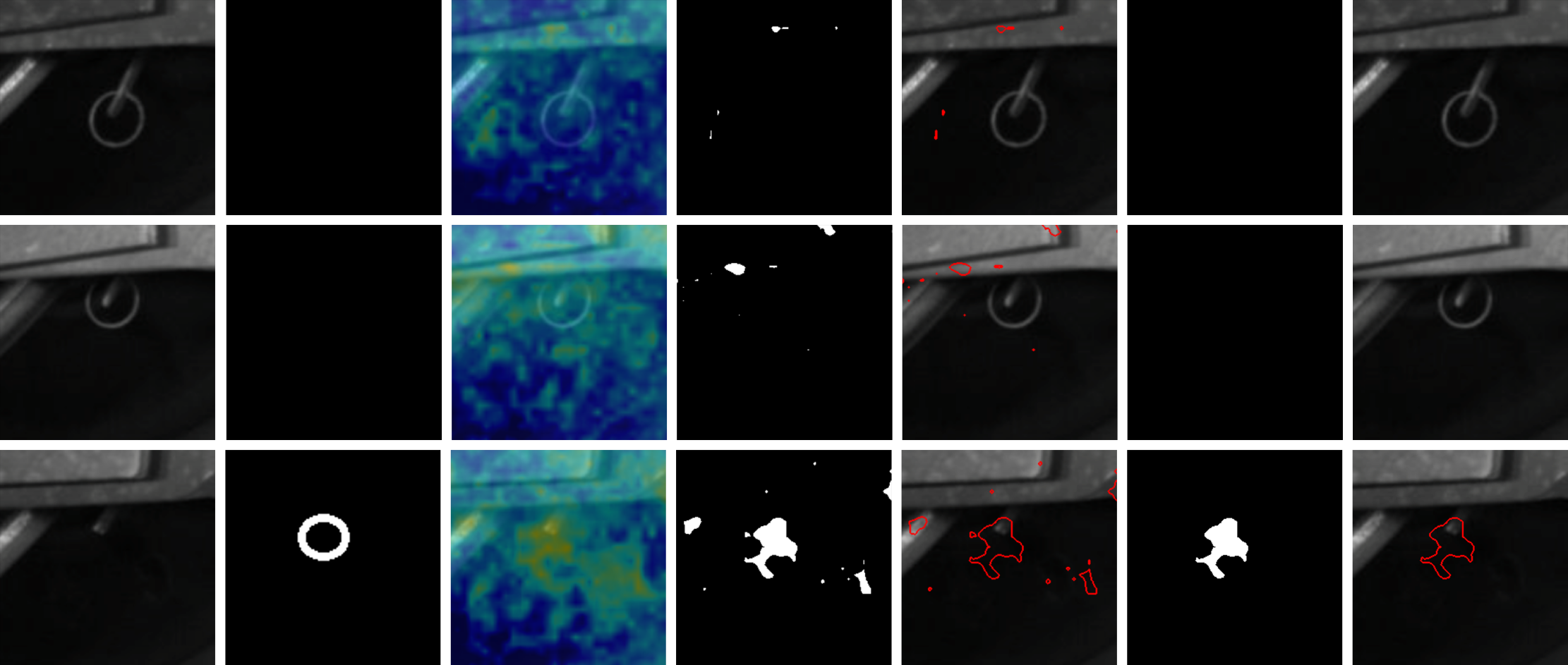}
\caption{Visualization of comparison experiments on round pin examination of freight train using Cflow. Column settings are identical to Fig. \ref{Fig: Fastflowwood}.}
\label{Fig: CflowTFDS}
\end{figure}

As shown in Fig. \ref{Fig: Fastflowwood} to Fig. \ref{Fig: CflowTFDS}, all top two rows are false alarms on anomaly-free images, and the proposed method succeeds in the recognition and elimination of false-alarm pixels. Although false alarms appear in heterogeneous distributions, the proposed method adaptively makes improvements. As presented in the last rows, defects in the test images reserve after filtering, which illustrates the discrimination ability of the post-precessing model.

The conclusion from visualization results conforms to the pixel-level performances in Table .\ref{tab: wood} and Table .\ref{tab: wood}. That is, the correct filtering of false alarms leads to the overall pixel-level metric enhancement than baseline detectors.

\subsection{Performance comparisons at image level}

Image level performance comparisons also confirm the positive effect of the proposed model.

In the MVTec-wood dataset, the image-level prediction scores of Cflow for anomaly-free images fall into the range $[0.07,0.59]$ while $[0.61,0.62]$ for defect images. That ranges of Fastflow are $[0.22,0.72]$ and $[0.75,0.79]$. The disjoint ranges result in perfect image-level AUROCs. Meanwhile, Cflow has two image-level false positives at 0.5876 and 0.5411, while fastflow gets one at 0.7196, which influences the F1-score. As the proposed method reduce image-level prediction scores according to the false-alarm elimination area, Fastflow and Cflow reach perfect F1-score after filtering.

Experiments on round pin examination of freight trains meet with more complex conditions than the former application leading to an overall performance degradation. However, the proposed model enhances both AUROCs and F1-scores on all comparison experiments using fuzzy knowledge of the joint distribution of scale and location.


\section{Conclusion}

In this Letter, we argue that the performance of OOD segmentation algorithms can be improved through a post-processing optimization method. An SVM classification model is proposed to learn the discriminative attributes of false alarms and defects in the specific application. Besides, an OOD segmentation model-dependent false-alarm sampling strategy is provided to generate training samples without additional labeled images. Experimental results of two SOTA anomaly detectors on two industrial applications verify the effect of the proposed method.


\end{document}